\documentclass[]{ailab}
\newcommand{\ours}{\textsc{EvoEnv}}

\newcommand{\VES}{verifiable environment synthesis}

\usepackage{wrapfig}
\usepackage{tabularx}
\usepackage{textcomp}
\usepackage{stfloats}
\usepackage{url}
\usepackage{verbatim}
\usepackage{titlesec}
\usepackage{tocloft}
\usepackage{adjustbox}
\usepackage{multirow}
\usepackage{pifont}
\usepackage{tikz}
\usepackage{comment}
\usepackage{amsmath,amssymb} 
\usepackage{colortbl}  
\usepackage{color}
\usepackage{booktabs} 
\usepackage{natbib} 
\setcitestyle{square,comma,numbers,sort&compress}
\usepackage{graphicx}    
\usepackage{subcaption} 
\usepackage{multirow} 
\usepackage{tcolorbox}
\usepackage{booktabs} 
\usepackage{subcaption} 
\usepackage[dvipsnames]{xcolor}
\usepackage{graphicx}
\usepackage{amsmath, amssymb}
\RequirePackage{xspace}
\makeatletter
\DeclareRobustCommand\onedot{\futurelet\@let@token\@onedot}
\def\@onedot{\ifx\@let@token.\else.\null\fi\xspace}

\usepackage{lineno}

\definecolor{darkblue}{rgb}{0, 0, 0.5}
\hypersetup{colorlinks=true, citecolor=darkblue, linkcolor=darkblue, urlcolor=darkblue}

\usepackage{wrapfig}  
\usepackage{lipsum}  

\newtcbtheorem[number within=section]{exmp}{Prompts}%
{breakable,colback=white!5!white,colframe=black!95!,fonttitle=\bfseries, left=.02in, right=.02in,bottom=.02in, top=.02in}{exmp}

\newtcbtheorem[number within=section]{case_data}{Examples}%
{breakable,
 colback=white!5!white,
 colframe={rgb,255:red,78; green,136; blue,114}, 
 fonttitle=\bfseries,
 left=.02in,
 right=.02in,
 bottom=.02in,
 top=.02in}{case_data}

\definecolor{mygreen}{HTML}{85A490}
\definecolor{myred}{HTML}{C75841}

\definecolor{cotred}{RGB}{200, 30, 30}
\definecolor{ForestGreen}{RGB}{34, 139, 34}

\usepackage[normalem]{ulem}

\makeatother

\definecolor{adptorange}{RGB}{248, 205, 172}
\definecolor{cmpblue}{RGB}{189, 215, 238}
\definecolor{cmpblue}{RGB}{189, 215, 238}

\definecolor{our_red}{RGB}{232,157,160}
\definecolor{our_blue}{RGB}{136,206,230}
\definecolor{our_orange}{RGB}{246,200,168}
\definecolor{our_green}{RGB}{178,211,164}

\definecolor{attn_code0}{RGB}{247,215,200}
\definecolor{attn_code1}{RGB}{238,169,139}
\definecolor{mlp_code0}{RGB}{204,201,221}
\definecolor{mlp_code1}{RGB}{102,95,153}

\definecolor{token_blue}{RGB}{84, 120, 140}

\usepackage{amsmath,amsfonts,bm}

\def\eqref#1{equation~\ref{#1}}

\def\1{\bm{1}}

\DeclareMathAlphabet{\mathsfit}{\encodingdefault}{\sfdefault}{m}{sl}
\SetMathAlphabet{\mathsfit}{bold}{\encodingdefault}{\sfdefault}{bx}{n}

\usepackage{amsmath,amsfonts,bm}

\def\eqref#1{equation~\ref{#1}}

\def\1{\bm{1}}

\DeclareMathAlphabet{\mathsfit}{\encodingdefault}{\sfdefault}{m}{sl}
\SetMathAlphabet{\mathsfit}{bold}{\encodingdefault}{\sfdefault}{bx}{n}

\usepackage{multirow}
\usepackage{diagbox}
\usepackage{makecell}
\usepackage{tabularx}
\usepackage{graphicx}

\usepackage{array}
\usepackage{rotating}

\definecolor{aliceblue}{rgb}{0.94, 0.97, 1.0}
\definecolor{citecolor}{HTML}{0071BC}
\definecolor{linkcolor}{HTML}{ED1C24}
\definecolor{darkgreen}{HTML}{539165}

\makeatletter
\newcommand{\thickhline}{%
 \noalign {\ifnum 0=`}\fi \hrule height 1pt
 \futurelet \reserved@a \@xhline
}
\makeatother

\usepackage{pifont}

\usepackage{pifont}       
\usepackage{bbding}       
\usepackage{fontawesome}
\usepackage{xspace}
\usepackage{float}
\usepackage{enumitem}

\newlength\savewidth

\newcolumntype{x}[1]{>{\centering\arraybackslash}p{#1pt}}
\newcolumntype{y}[1]{>{\raggedright\arraybackslash}p{#1pt}}
\newcolumntype{z}[1]{>{\raggedleft\arraybackslash}p{#1pt}}

\renewcommand{\paragraph}[1]{\vspace{1mm}\noindent\textbf{#1}}
\usepackage{colortbl}
\usepackage{xcolor}
\usepackage{wrapfig}

\renewcommand{\paragraph}[1]{\vspace{1.25mm}\noindent\textbf{#1}}

\usepackage{algorithm}
\usepackage{listings}

\definecolor{codeblue}{rgb}{0.25, 0.5, 0.5}
\definecolor{codekw}{rgb}{0.35, 0.35, 0.75}
\lstdefinestyle{Pytorch}{
    language = Python,
    backgroundcolor = \color{white},
    basicstyle = \fontsize{9pt}{8pt}\selectfont\ttfamily\bfseries,
    columns = fullflexible,
    aboveskip=1pt,
    belowskip=1pt,
    breaklines = true,
    captionpos = b,
    commentstyle = \color{codeblue},
    keywordstyle = \color{codekw},
}

\definecolor{colSubject}{HTML}{D32F2F}   
\definecolor{colAction}{HTML}{F57C00}    
\definecolor{colDetail}{HTML}{388E3C}    
\definecolor{colSpatial}{HTML}{1976D2}   
\definecolor{colMood}{HTML}{7B1FA2}      
\definecolor{colKnow}{HTML}{AFB42B}      

\usepackage{xcolor}

\usepackage{tikz}
\usetikzlibrary{tikzmark, calc, shadows.blur, shapes.geometric, fit, positioning}
\usepackage{xparse}
\pgfdeclarelayer{background}
\pgfdeclarelayer{floatbox}
\pgfdeclarelayer{foreground}
\pgfsetlayers{background,floatbox,main,foreground}

\newcounter{hcellcount}

\NewDocumentCommand{\hctext}{m}{\csname hctext@#1\endcsname}
\NewDocumentCommand{\sethctext}{mm}{\expandafter\gdef\csname hctext@#1\endcsname{#2}}

\definecolor{scoreRed}{RGB}{200, 0, 0}
\definecolor{grayText}{RGB}{120, 120, 120}

\definecolor{green}{HTML}{009000}
\definecolor{red}{HTML}{ea4335}

\definecolor{cvblue}{rgb}{0.15, 0.45, 0.68}

\usepackage{mathtools}
\usepackage{nicefrac}
\usepackage{microtype}
\usepackage{algpseudocode}
\usepackage{longtable}
\usepackage{ragged2e}

\definecolor{codebg}{HTML}{F7F8FA}
\definecolor{codeframe}{HTML}{D0D7DE}

\usepackage[most]{tcolorbox}
\tcbuselibrary{listings,skins,breakable}

\usepackage[scaled=0.85]{beramono}

\definecolor{codebg}{HTML}{FAFBFC}      
\definecolor{codeframe}{HTML}{E1E4E8}   
\definecolor{codekw}{HTML}{0033B3}      
\definecolor{codecmt}{HTML}{8A8F98}     
\definecolor{codestr}{HTML}{067D17}     
\definecolor{codenum}{HTML}{B0B6BC}     

\newtcblisting{pycodebox}{
  listing engine=listings,
  listing options={
    language=Python,
    basicstyle=\scriptsize\ttfamily,
    keywordstyle=\color{codekw},
    commentstyle=\color{codecmt}\itshape,
    stringstyle=\color{codestr},
    numbers=left,
    numberstyle=\tiny\color{codenum},
    numbersep=8pt,
    xleftmargin=16pt,
    breaklines=true,
    breakatwhitespace=false,
    showstringspaces=false,
    tabsize=4,
    columns=fullflexible,
    keepspaces=true,
    upquote=true,
    aboveskip=2pt,
    belowskip=2pt,
  },
  listing only,
  enhanced,
  colback=codebg,
  colframe=codeframe,
  boxrule=0.4pt,
  arc=3pt,
  left=1mm,
  right=2mm,
  top=1.5mm,
  bottom=1.5mm,
  boxsep=0pt
}

\providecommand{\indic}{\mathbb{I}}

\title{\centering Learning to Build the Environment: Self-Evolving Reasoning RL via Verifiable Environment Synthesis}

\author[1]{Yucheng Shi}
\author[1]{Zhenwen Liang}
\author[1]{Kishan Panaganti}
\author[1]{Dian Yu}
\author[1]{Wenhao Yu}
\author[1]{Haitao Mi}

\affiliation[1]{Tencent HY LLM}

\abstract{
We pursue a vision for self-improving language models in which the model does not merely generate problems or traces to imitate, but constructs the environments that train it. In zero-data reasoning RL, this reframes self-improvement from a data-generation loop into an environment-construction loop, where each artifact is a reusable executable object that samples instances, computes references, and scores responses. Whether this vision sustains improvement hinges on a single property: the environments must exhibit stable solve--verify asymmetry: the model must be able to write an oracle once that it cannot reliably execute in natural language on fresh instances. This asymmetry takes two complementary forms. Some tasks are algorithmically hard to reason through but trivial as code: a dynamic program or graph traversal, compiled once, yields unboundedly many calibrated instances. Others are intrinsically hard to solve but easy to verify, like planted subset-sum or constraint satisfaction. Both create a durable gap between proposing and solving that the policy cannot close by gaming the verifier, and it is this gap that keeps reward informative as the learner improves. We instantiate this view in \ours{}, a single-policy generator-solver method that synthesizes Python environments from ten seeds and admits them only after staged validation, semantic self-review, solver-relative difficulty calibration, and novelty checks. The strongest evidence comes from the already-strong regime: on Qwen3-4B-Thinking, fixed public-data RLVR and fixed hand-crafted environment RLVR reduce the average, while \ours{} improves it from $72.4$ to $74.8$, a relative gain of $3.3\%$. Stable self-improvement, we suggest, depends not on producing more synthetic data, but on models learning to construct worlds whose difficulty stays structurally beyond their own reach.
}

\date{\today}

\begin{document}
\thispagestyle{firstheader}
\maketitle

\section{Introduction}
\label{sec:intro}

Language-model self-improvement is often framed as a data-generation problem: the model produces more questions, traces, solutions, or hard examples near its current frontier~\cite{rzero, spiral, liu2025spice}. This framing is useful but incomplete. Capable LLMs do not improve only by imitating additional examples; they improve by interacting with environments that generate situations, impose constraints, and return feedback through execution, tools, tests, or state changes~\cite{zeng2026glm, team2026kimi}. We ask whether a language model can learn not only from self-generated examples, but from self-constructed training environments.
We study this question in a deliberately controlled setting: zero-data reinforcement learning with verifiable rewards (RLVR) for reasoning. A reasoning environment is a reusable executable artifact with four routines: a sampler that generates latent task instances, an oracle that computes reference answers, a renderer that turns instances into natural-language prompts, and a scorer that evaluates solver responses. The model may author this artifact, but once the artifact is validated and admitted to the pool, its rewards are determined by execution rather than by the model's current sampled answers.

This distinction addresses two limitations of existing self-training recipes. Standard RLVR obtains reliable supervision from fixed datasets, answer-equivalence rules, unit tests, or executable checkers~\citep{deepseekr1,dapo,prorl}. However, the verified distribution is fixed. As the policy improves, many prompts become almost always solved, others remain almost always failed, and group-relative methods lose useful reward variation. Continued optimization on a saturated pool can then narrow behavior or cause forgetting rather than expand capability~\citep{he2026urlvr,wu2025mirage,shao2025spurious}. Stable verifiers are therefore not sufficient; the verified distribution must also stay near the solver's changing frontier.

Per-instance self-play provides adaptivity but can compromise label stability. If new prompts are labeled by self-consistency, majority vote, semantic clustering, or another signal derived from the same policy being optimized, the reward function moves with the learner~\cite{rzero, spiral, liu2025spice}. If the model instead generates a one-off executable verifier for a single problem, correctness is better grounded, but the artifact is consumed after one rollout~\cite{absolutezero}. The resulting system still lacks a durable object that can be validated once, sampled repeatedly, calibrated against the current solver, retired when it saturates, and later reused as seed material.

Our thesis is that the unit of self-synthesis should be the environment rather than the individual problem. A generated problem gives one prompt and one label; a generated environment gives a distribution of prompts and a reusable executable reward source. The reason this environment-level unit is tractable is not merely amortization across rollouts, but a structural property we call \emph{stable solve--verify asymmetry}: across a wide class of reasoning tasks, authoring or checking an executable procedure is easier than carrying out the corresponding natural-language reasoning process on fresh instances.

This asymmetry appears in two complementary forms. In \emph{algorithmic} tasks, such as dynamic programming, graph traversal, modular recurrence, sorting, and sequence computation, the model may be able to write a compact oracle even when it cannot reliably execute that algorithm in natural language on arbitrary rendered instances. In \emph{verification} tasks, such as planted subset-sum, feasibility checking, and constraint satisfaction, producing a valid answer can be hard, while checking a proposed answer is simple. Both forms create a durable gap between proposing and solving, a gap that the policy cannot close by gaming the verifier, because the verifier is frozen code. It is this gap that keeps reward informative as the learner improves, and it is what distinguishes self-\emph{built environments} from self-\emph{generated problems}.

We instantiate this view in \ours{}, a single-policy dual-role trainer. The same policy alternates between a \emph{generator} role, which proposes Python environments from a small seed pool, and a \emph{solver} role, which answers fresh prompts sampled from the accepted environment pool. Candidate environments enter the pool only when they satisfy four contracts: they execute under a strict interface; their oracle and scorer match the advertised task under conservative semantic self-review; sampled instances are hard-but-solvable for the current policy; and the pool remains broad enough to avoid template collapse. We enforce these contracts through L1--L5 validation, semantic self-review, novelty gating, in-batch deduplication, and pool rotation.

The strongest evidence for this framing comes from the already-strong model regime.  On Qwen3-4B-Thinking, fixed public-data RLVR and fixed hand-crafted environment RLVR both reduce the average score in Table~\ref{tab:main}, while \ours{} improves it.  This result suggests that self-evolving environments are not merely a way to obtain more synthetic data for weak models.  They are a way to keep reward stable and frontier-calibrated when static distributions have already saturated or become misaligned with the learner. Our contributions are:

\begin{enumerate}[leftmargin=1.35em,itemsep=0.15em,topsep=0.25em]
    \item We formulate \VES{} for zero-data reasoning RL, identifying \emph{stable solve--verify asymmetry}, in both algorithmic and verification forms, as the structural property that allows self-built environments to serve as durable reward sources rather than policy-coupled pseudo-labels.
    \item We introduce \ours{}, a single-policy proposer--solver algorithm that curates a self-generated environment pool through staged validation, semantic self-review, solver-relative difficulty calibration, novelty control, and pool rotation.
    \item We show that \ours{} improves three model families and, in particular, improves an already-strong thinking-mode checkpoint where both fixed public-data RLVR and fixed hand-crafted environment RLVR reduce average performance.
\end{enumerate}

\section{Related Work}
\label{sec:related}

We position \ours{} by asking two questions: what object is reused during training, and where does its reward come from?  Prior work provides stable verifiers, adaptive self-generated curricula, executable task checks, or environment-level training.  \ours{} combines these ingredients in a specific setting: zero-data reasoning RL, where the learner itself authors reusable executable environments, and each environment is admitted only after validation, solver-relative calibration, and novelty filtering.

\paragraph{Fixed-distribution RLVR.}
Verifier-backed RL has been highly effective for reasoning because outcome rewards can be computed without a learned preference model~\citep{deepseekr1,dapo,prorl}.  Most RLVR systems, however, train on a fixed set of prompts and checkers.  This is reliable but not adaptive: once examples become almost always solved or almost always failed, group-relative rewards lose useful variation, and continued optimization can lead to saturation, over-specialization, or forgetting~\citep{he2026urlvr,wu2025mirage,shao2025spurious}.  \ours{} keeps the key benefit of RLVR---execution-grounded reward---but replaces the static prompt pool with a changing pool of validated executable environments.

\paragraph{Self-generated curricula.}
A broad line of self-improvement methods trains on model-generated rationales, questions, preferences, or judgments~\citep{zelikman2022star,singh2024restem,chen2024spin,yuan2024selfrewarding,hosseini2024vstar}.  Recent zero-data methods go further by letting the model propose its own tasks and estimate correctness through majority vote, self-consistency, internal feedback, or co-evolving agents~\citep{rzero,fang2025serl,chen2025selfquestioning,ttrl,empo,scpo,zhao2025intuitor,kwan2025opensir,zhang2025pasodoble,wang2025socraticzero,chen2025multiagentevolve}.  These approaches are adaptive, but their reward signals are often policy-coupled: the same model family that learns also helps decide what is correct.  Novelty-based variants such as EVOL-RL mitigate collapse in these loops~\citep{zhou2025evolrl}.  \ours{} shares the goal of adaptive self-improvement, but it does not trust the model's sampled answers as labels; it trusts only executable artifacts that pass admission and are then frozen for scoring.

\paragraph{Executable grounding.}
Several methods ground self-generated tasks through code or tests.  Absolute Zero Reasoner verifies program/input/output tasks by execution~\citep{absolutezero}; Self-Challenging Agents generate Code-as-Task verifiers~\citep{zhou2025sca}; SPC trains adversarial critics for reasoning errors~\citep{chen2025spc}; and agentic systems extend similar ideas to tool use and software engineering~\citep{xia2025agent0,zhu2025sweplayground,wei2025ssr}.  These works are closer to ours because correctness is no longer purely self-believed.  The main difference is the reuse unit: they typically generate tasks, tests, or episodes, whereas \ours{} generates an environment-level object that can sample many fresh instances, be calibrated against the current solver, retired when saturated, and reused as seed material.

\paragraph{Environment-level training.}
The closest neighbors also treat environments as training objects.  Some environments are externally given, such as fixed-rule games, language games, document corpora, or hand-authored verifiable Python suites~\citep{spiral,liu2025selfredteam,kuba2025lsp,lu2025searchselfplay,liu2025spice,rlve}.  Others synthesize tool, web, UI, or embodied environments through offline pipelines~\citep{song2026envscaler,wu2026autowebworld,zhang2026infiniteweb,chae2026verienv,ramrakhya2025autoplay,lei2025embomatrix}.  Learned-simulator approaches instead use an LLM world model, so reward depends on the simulator's beliefs~\citep{chen2025dreamgym,wang2025uisimulator,fang2025webevolver}.  \ours{} studies a more controlled case: the same policy that solves reasoning tasks also authors deterministic Python environments, while admission is on-policy, solver-calibrated, and execution-grounded rather than human-authored, offline-pipeline-generated, or simulator-defined.  Appendix~\ref{app:positioning} gives a fuller comparison.

\section{Method}
\label{sec:method}

\begin{figure}[t]
\centering
\includegraphics[width=\textwidth]{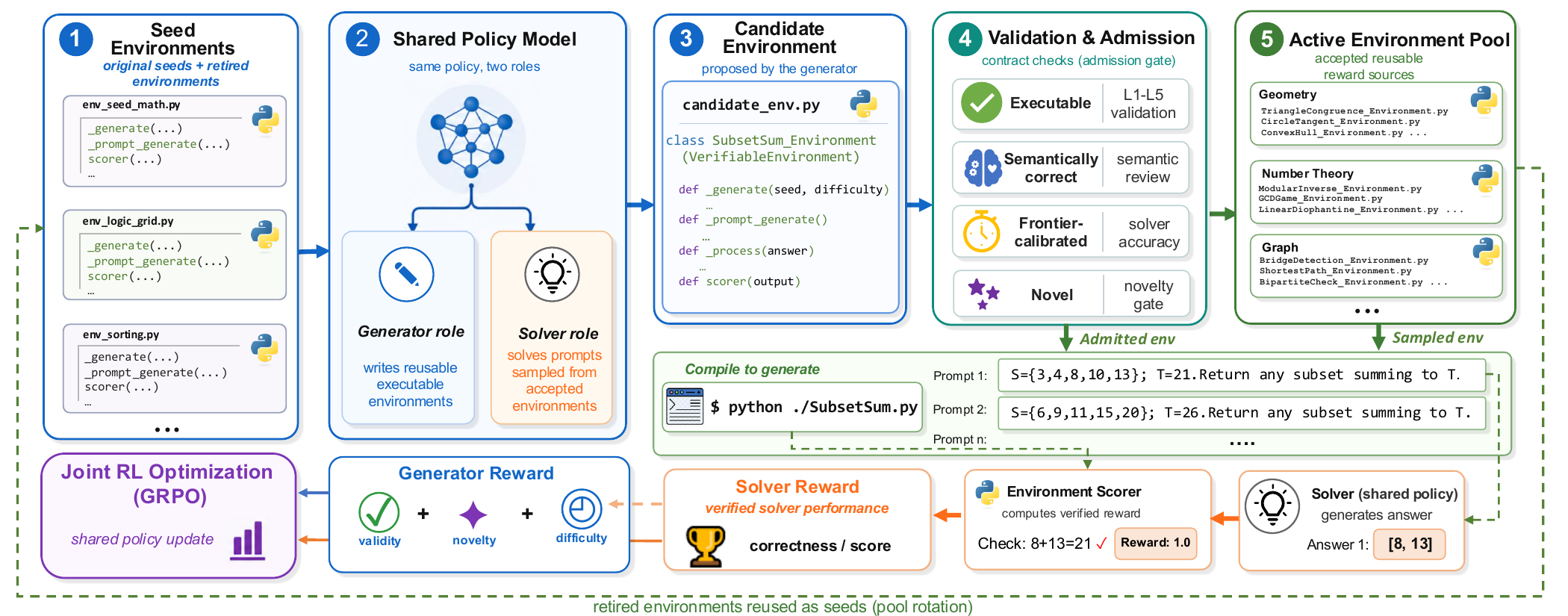}
\vspace{-10pt}
\caption{Overview of \ours{}.  A single policy generates reusable executable environments and solves fresh instances sampled from the accepted pool.  Candidate environments enter the pool only after validation, semantic review, solver-relative difficulty calibration, and novelty filtering.}
\label{fig:overview}
\vspace{-10pt}
\end{figure}

\ours{} trains a single policy to do two related things: solve verifiable reasoning tasks, and construct the executable environments from which such tasks are sampled. The central object is therefore not a generated problem, but a reusable environment. A candidate environment is allowed to influence solver training only after it passes mechanical validation, conservative semantic review, solver-relative difficulty calibration, and novelty filtering.  The design goal is simple: teach the model to construct new training worlds, while ensuring that rewards used for solver updates come from frozen execution rather than from the model's current sampled answers. A complete workflow is presented in Figure~\ref{fig:overview} and Algorithm~\ref{alg:evoenv}.

\subsection{Environment interface}
\label{sec:method:env}

\begin{wrapfigure}{r}{0.46\linewidth}
\vspace{-8pt}
\begin{pycodebox}
import random

class SortingEnv(VerifiableEnvironment):

    def _generate(self, seed, difficulty):   # G_e
        rng = random.Random(seed)
        z = [rng.randint(0, 99)
             for _ in range(difficulty)]
        return z, sorted(z)                  # (z, a)

    def _prompt_generate(self, z):           # Pi_e
        arr = " ".join(map(str, z))
        return "Sort: " + arr                # x

    def _process(self, y):
        try:
            return list(map(int, y.split()))
        except ValueError:
            return None

    def scorer(self, y, z, a):               # S_e
        return float(self._process(y) == a)
\end{pycodebox}
\vspace{-8pt}
\caption{A minimal sorting environment. The method \texttt{\_generate} implements $G_e$, \texttt{\_prompt\_generate} implements $\Pi_e$, and \texttt{\_process} with \texttt{scorer} implements $S_e$.}
\label{fig:sorting_code}
\vspace{-10pt}
\end{wrapfigure}

A verifiable environment is a reusable executable object, not a single labeled problem. We write it as $e = (G_e,\Pi_e,S_e)$.  Given a seed $\sigma$ and difficulty parameter $\delta$, the generator--oracle routine produces a latent instance and reference answer,
$
(z,a)=G_e(\sigma,\delta).
$
The solver observes only the rendered prompt $x=\Pi_e(z)$ and receives reward
$
R_e(y;\sigma,\delta)=S_e(y,z,a)
$
for its response $y$.  Thus, once an environment is admitted, the reward source is a frozen executable path: for fixed $e,\sigma,\delta$, the current policy can change only the sampled response, not the reference answer or scoring rule.

Figure~\ref{fig:sorting_code} instantiates this interface with a minimal sorting environment. The environment samples an array, computes the executable reference answer with \texttt{sorted}, renders the array as a model-facing prompt, and scores the parsed response by exact comparison.  This illustrates the operational solve--verify asymmetry used throughout \ours{}: writing and running this code is relatively easy, but solving many fresh rendered instances still supplies nontrivial training signal for the language model.  

In implementation, each candidate is emitted as a Python subclass of \texttt{VerifiableEnvironment}.  The method \texttt{\_generate(seed, difficulty)} implements $G_e$ by constructing $z$ and computing $a$; \texttt{\_prompt\_generate} implements $\Pi_e$; and \texttt{\_process} together with \texttt{scorer} implements $S_e$.  Candidates are restricted to an approved standard-library subset and executed in sandboxed subprocesses with wall-clock timeouts.

This interface covers both equality-check oracles, such as sorting, dynamic programming, graph traversal, modular recurrence, and sequence computation, and feasibility-check oracles, such as planted subset-sum and constraint satisfaction.  Feasibility tasks require stronger scorer probes because multiple answers may be valid.  Appendix \ref{app:env_anatomy} gives the full data-flow diagram, and Appendix \ref{app:subset_example} provides a complete planted subset-sum example.

\subsection{Validation and semantic review}
\label{sec:method:validate}

Our method is equipped with multi-layer validation. Let $\ell(e)$ denote the highest validation layer reached by candidate $e$. L1 extracts parseable Python and checks that the expected class and methods exist. L2 instantiates the class and runs generation, prompt rendering, parsing, and scoring on several seeds and difficulty settings.  L3 checks determinism by repeating generation under identical seeds and comparing the latent instance, prompt, and reference object.  L4 checks non-triviality by requiring variation across seeds and difficulty values.  L5 checks the local scorer contract: the stored reference object must score positively; injected perturbations, malformed answers, and type-mismatched answers must not; and parsing must not leak the hidden reference object.  Only L5 candidates proceed to semantic review and solver-relative calibration. 

Mechanical execution alone cannot prove that the generated code implements the task described in the prompt.  A candidate can be deterministic and non-trivial while computing the wrong recurrence, rewarding the wrong target, or accepting malformed answers.  We therefore add a conservative semantic-review filter. The reviewer receives the candidate source code, one or more concrete generated instances, the reference object, the rendered prompt, and scorer probes.  It is asked to perform a code-review task: trace the advertised task, check that the generator computes the advertised quantity, and search for domain relabeling, hidden answer leakage, and overly permissive parsing. 

The reviewer is the same policy used elsewhere in training, but its verdict is not used as generator reward.  This distinction is important.  The generator's scalar reward is computed from mechanical validation, solver-relative difficulty, and novelty; the semantic reviewer only decides whether an otherwise valid candidate may enter the active solver-training pool.  A rejected candidate may still contribute its mechanically computed generator rollout reward, but it cannot become a reward source for solver training.  This removes the direct channel by which the generator could be optimized to satisfy the reviewer's natural-language preferences.

We run $K_{\rm rev}=3$ independent reviews and use an any-reject rule: if any review identifies a likely semantic bug, the candidate is rejected from pool admission.  The review task is substantially more local than solving benchmark problems from scratch: the reviewer sees the source, hidden state, reference object, and scorer behavior, and only needs to check consistency among them.  As an additional sanity check, we audit this same-policy review against a stronger external reviewer and find high agreement; details are reported in Appendix~\ref{app:semantic_audit}.  

\subsection{Difficulty and novelty rewards}
\label{sec:method:reward}

For each L5 candidate, we estimate whether its sampled instances produce useful outcome variation for the current solver.  We sample $m=8$ calibration instances by running $G_e$, draw a single solver response per instance, and average:
\begin{equation}
\label{eq:empirical_accuracy}
\hat a_m(e;\pi_\theta)
=
\frac{1}{m}
\sum_{i=1}^{m}
\mathbb I\!\left[
S_e(y_i,z_i,a_i)>0
\right],
\qquad
(z_i,a_i)=G_e(\sigma_i,\delta),
\quad
y_i\sim \pi_\theta(\cdot\mid \Pi_e(z_i)).
\end{equation}
We use one response per instance to estimate the pass rate without inflating compute.  Candidates with $\hat a_m(e;\pi_\theta)=0$ are too hard, underspecified, or overly strict under the current solver; candidates with $\hat a_m(e;\pi_\theta)=1$ are saturated or overly permissive.  We therefore require $0<\hat a_m(e;\pi_\theta)<1$ for admission.
The generator's difficulty reward grades each L5 candidate by its solver-relative pass rate using a piecewise schedule:
\begin{equation}
\label{eq:qunc}
Q_{\rm unc}(e;\theta)
=
\exp\!\left(
-\frac{
\left(\hat a_m(e;\pi_\theta)-a^\star\right)^2
}{
2\sigma_a^2
}
\right),
\qquad
a^\star=0.3.
\end{equation}
The target $a^\star=0.3$ biases generation toward environments that are solvable but still challenging: the current solver succeeds often enough to produce positive examples, but fails often enough to preserve useful reward variation.  We choose a target below $0.5$ because near-half accuracy candidates can become saturated quickly as the solver improves.

Novelty prevents the generator from repeatedly producing the first template that passes validation.  We embed each environment with a frozen external embedding model, \texttt{all-MiniLM-L6-v2}~\cite{reimers2019sentence}, rather than with the training policy itself.  Each candidate has two embeddings: a prompt embedding $u_e$ from its \texttt{prompt\_template} and a code embedding $v_e$ from its cleaned \texttt{\_generate} body.  Let $\mathcal C_p$ and $\mathcal C_c$ be caches of prompt and code embeddings for previously admitted environments.  We compute
\begin{equation}
\label{eq:novelty_sim}
\mathrm{sim}_t(e)
=
\lambda
\max_{u'\in\mathcal C_p}\cos(u_e,u')
+
(1-\lambda)
\max_{v'\in\mathcal C_c}\cos(v_e,v'),
\qquad
\lambda=0.5,
\end{equation}
and define
$
N_t(e)=1-\mathrm{sim}_t(e).
$
The two-view novelty score is a guardrail to prevent surface-level duplicates. Prompt embeddings alone can miss code clones with different story wrappers, and code embeddings alone can miss the same task written in different language.  Combining prompt and code views makes exact or near-exact duplication less attractive.  At the same time, we also acknowledge that surface variants can still be useful training environments when they are semantically valid and present the solver with different natural-language contexts.
The novelty weight adapts to the repetitiveness of the accepted stream.  Let $\bar{s}_t$ be an exponential moving average of within-batch maximum similarity.  We set
\begin{equation}
\label{eq:gamma_schedule}
\gamma_t
=
\gamma_{\min}
+
(\gamma_{\max}-\gamma_{\min})
\cdot
\mathrm{clip}
\left(
\frac{\bar{s}_t-\tau_{\mathrm{low}}}
{\tau_{\mathrm{high}}-\tau_{\mathrm{low}}},
0,1
\right).
\end{equation}
Exploration pressure therefore rises when the accepted pool becomes repetitive and relaxes when new environments are already diverse.
The full generator reward combines layered validation with a novelty bonus:
\begin{equation}
\label{eq:rgen}
R_{\rm gen}(e;\theta,t)
=
Q_{\rm val}\!\left(\ell(e),\hat a_m(e;\pi_\theta)\right)
+
\mathbb I[\ell(e)\ge 2]\,\gamma_t\,N_t(e).
\end{equation}
Here the validation term is
\begin{equation}
\label{eq:qval}
Q_{\rm val}(\ell,\hat a)
=
-\mathbb I[\ell<1]
-\frac{1}{2}\mathbb I[\ell=1]
-\frac{1}{4}\mathbb I[\ell=2]
+\mathbb I[\ell=5]\,Q_{\rm unc}(\hat a).
\end{equation}
The validation term penalizes mechanically invalid candidates, assigns zero reward to candidates that pass execution-level checks but fail top-layer validation, and gives the solver-relative uncertainty reward $Q_{\rm unc}$ only to L5 candidates.  The novelty bonus is gated by $\ell(e)\ge2$, so unparseable or syntactically broken candidates cannot recover through novelty alone.  Crucially, the semantic-review
verdict is not a term in $R_{\rm gen}$; it only affects the pool-admission decision.

\subsection{Pool admission and joint optimization}
\label{sec:method:grpo}

A candidate is inserted into the active solver-training pool only if it satisfies all four admission requirements:
\begin{equation}
\label{eq:accept}
A_t(e)=
\indic[\ell(e)=5]\cdot
\indic[\mathrm{review}(e)=1]\cdot
\indic[0<\hat a_m(e;\pi_\theta)<1]\cdot
\indic[\mathrm{sim}_t(e)<\tau_{\mathrm{gate}}].
\end{equation}
Accepted environments enter $\mathcal P_t$ and may later be used both for solver training and as examples in $\mathcal S_t$ for future generator prompts.
At each training step, we form solver and generator rollout groups.  

A solver group samples an environment $e\sim\mathcal P_t$, generates an instance and prompt using $G_e$ and $\Pi_e$, samples multiple responses from $\pi_\theta$, and scores each
response with the frozen environment scorer $S_e$.  Thus the solver receives ordinary verifier-backed RL feedback: it is rewarded only for producing an answer accepted by the environment.

A generator group samples multiple candidate environments from the same few-shot generator prompt built from $\mathcal S_t$.  Each candidate receives the reward in Eq.~\eqref{eq:rgen}; candidates satisfying Eq.~\eqref{eq:accept} are additionally admitted into the active pool.  We normalize advantages within solver groups and generator groups separately, then optimize the shared policy with a role-conditioned GRPO objective:
\begin{equation}
\label{eq:joint_loss}
\mathcal L(\theta)
=
\mathcal L_{\rm solver}^{\rm GRPO}(\theta)
+
w_{\rm gen}\mathcal L_{\rm gen}^{\rm GRPO}(\theta)
+
\beta\,\mathrm{KL}\!\left(\pi_\theta \,\|\, \pi_{\rm ref}\right),
\qquad
w_{\rm gen}=0.3.
\end{equation}
The two roles use distinct prompts but share parameters.  The single-policy design is
chosen for efficiency: the same model that benefits from executable reasoning
environments also learns the code patterns needed to construct them.  In practice,
separate group normalization and the generator weight $w_{\rm gen}$ prevent long,
high-variance generator trajectories from dominating solver updates.

Pool rotation controls the long-horizon curriculum.  After an environment has been used
for a fixed number of solver epochs, it is retired from $\mathcal P_t$ and is no longer
sampled for direct solver training.  Retired environments can instead enter
$\mathcal S_t$ as few-shot examples for future generator prompts.  Original seed
environments are protected by a floor $\rho_{\min}=0.2$: an original seed is not retired
if doing so would reduce the share of originals in the active pool below the floor.
This keeps the generator context anchored while allowing the solver-training pool to
move toward newly synthesized environments.


\begin{algorithm}[t]
\caption{One self-evolution iteration of \ours{}.}
\label{alg:evoenv}
\small
\begin{algorithmic}[1]
\Require policy $\pi_{\theta_t}$; active pool $\mathcal P_t$; seed pool $\mathcal S_t$; novelty cache $\mathcal C_t$
\Ensure updated policy $\pi_{\theta_{t+1}}$ and pools $(\mathcal P_{t+1},\mathcal S_{t+1})$
\State Initialize generator data $\mathcal D_t^g\gets\emptyset$, solver data $\mathcal D_t^s\gets\emptyset$, accepted set $\mathcal A_t\gets\emptyset$
\State Sample candidate environments $\{e_i,\tau_i^g\}_{i=1}^M \sim \pi_{\theta_t}(\cdot\mid \mathsf{Prompt}_{\rm gen}(\mathcal S_t))$
\For{$i=1,\ldots,M$}
    \State $(r_i^g,A_i)\gets \textsc{EvalEnv}(e_i;\pi_{\theta_t},\mathcal C_t)$
    \State $\mathcal D_t^g\gets \mathcal D_t^g\cup\{(\tau_i^g,r_i^g)\}$; \quad if $A_i=1$, add $e_i$ to $\mathcal A_t$
\EndFor
\For{$j=1,\ldots,B$}
    \State Sample $e\sim \mathcal P_t\cup\mathcal A_t$, $(z,a)\sim G_e$, $x=\Pi_e(z)$, and responses $\{y_{jm}\}_{m=1}^K\sim\pi_{\theta_t}(\cdot\mid x)$
    \State $\mathcal D_t^s\gets \mathcal D_t^s\cup\{(\tau_{jm}^s,S_e(y_{jm},z,a))\}_{m=1}^K$
\EndFor
\State $\theta_{t+1}\gets \textsc{GRPO}(\theta_t;\mathcal D_t^g,\mathcal D_t^s)$
\State $(\mathcal P_{t+1},\mathcal S_{t+1},\mathcal C_{t+1})\gets \textsc{UpdatePool}(\mathcal P_t,\mathcal S_t,\mathcal C_t,\mathcal A_t)$
\end{algorithmic}
\end{algorithm}

\section{Experiments}
\label{sec:experiments}

Our experiments test three questions.  (Q1) Does \ours{} improve downstream reasoning over the untrained base and strong RLVR baselines?  (Q2) Does environment synthesis sustain a frontier signal over training, or does it saturate like a fixed pool?  (Q3) Which components are load-bearing?  

\subsection{Setup}
\label{sec:exp:setup}

\paragraph{Models.}
We evaluate on three diverse base models: \textbf{Qwen3-4B-Instruct-2507}~\cite{qwen3technicalreport},
a standard instruction-tuned 4B model; \textbf{Qwen3-4B-Thinking-2507}~\cite{qwen3technicalreport},
an already-strong thinking-mode checkpoint; and
\textbf{Nemotron-Cascade-8B-IFRL}~\cite{yang2025nemotroncascade}, an 8B thinking model from a different family.

\paragraph{Seeds and training.}
All runs start from the same fixed set of ten seed environments spanning sorting, dynamic programming, graph, and number-theoretic templates (Appendix~\ref{app:seeds}).  No external problem-answer data is used.  We train with single-policy GRPO, generator advantage scale $w_{\mathrm{gen}}=0.3$, and the hyperparameters in Appendix~\ref{app:hparams}.  All models are trained for $100$ steps.

\paragraph{Evaluation.}
We report pass@1 [\%] across eight benchmarks using the Nemo-Skills toolkit~\cite{moshkov2025aimo2}.
For the small-problem-set competition math benchmarks, AIME 2024, AIME 2025, HMMT Feb 2025,
Beyond-AIME~\cite{seed2025thinking}, and Brumo 2025~\cite{balunovic2025matharena},
we average over 16 runs to reduce variance; AMC 2023~\cite{balunovic2025matharena},
GPQA Diamond~\cite{rein2023gpqa}, and LiveCodeBench v6, 2024-08--2025-05 (LCB)~\cite{jain2024livecodebench}
are reported as single-run pass@1. Unless stated otherwise, evaluation uses up to $128\mathrm{k}$ context, symbolic-equivalence checking, temperature $0.6$, and top-$p$ $0.95$.

\paragraph{Baselines.}
We compare against: \textbf{Untrained}, the base checkpoint under the same decoding protocol; \textbf{R-Zero}~\citep{rzero}, a zero-data proposer--solver RLVR method; \textbf{DAPO}~\citep{dapo}, GRPO trained on DAPO-Math-17k at matched compute; and \textbf{RLVE}~\citep{rlve}, fixed-environment RLVR with hand-curated environments and matched training steps. 

\subsection{Main results: adaptive environments remain useful where fixed distributions fail}
\label{sec:exp:main}

\begin{table}[t]
\centering
\caption{Main results across three base models and eight benchmarks.  ``Avg'' is the mean result across the reported cells in that row.  Best result per model block is \textbf{bolded}.}
\label{tab:main}
\setlength{\tabcolsep}{3.2pt}
\footnotesize
\resizebox{\textwidth}{!}{%
\begin{tabular}{@{}ll cccccccc c@{}}
\toprule
\textbf{Base Model} & \textbf{Method} & \textbf{AIME'24} & \textbf{AIME'25} & \textbf{AMC'23} & \textbf{HMMT} & \textbf{B-AIME} & \textbf{Brumo} & \textbf{GPQA} & \textbf{LCB} & \textbf{Avg}\\
\midrule
\multirow{5}{*}{\shortstack[l]{Qwen3-4B\\Instruct-2507}}
  & Untrained          & 58.8 & 44.0 & 87.5 & 30.0 & 32.0 & 56.5 & 52.5 & 32.6 & 49.2\\
  & R-Zero             & 55.0 & 43.3 & 87.5 & 26.5 & 27.0 & 55.2 & 54.0 & 33.5 & 47.8\\
  & DAPO               & 50.0 & 44.8 & 92.5 & 19.8 & 22.2 & 47.1 & 56.1 & 33.5 & 45.8 \\
  & RLVE               & 57.1 & 44.3 & 92.5 & 27.7 & 32.8 & 57.9 & \textbf{57.1} & 32.6 & 50.3 \\
  & \textbf{\ours}     & \textbf{61.9} & \textbf{52.5} & \textbf{95.0} & \textbf{30.0} & \textbf{33.6} & \textbf{61.2} & 55.0 & \textbf{35.5} & \textbf{53.1} \\
\midrule
\multirow{5}{*}{\shortstack[l]{Qwen3-4B\\Thinking-2507}}
  & Untrained          & \textbf{86.7} & 78.8 & 97.5 & 56.9 & 53.1 & \textbf{81.9} & 65.2 & 59.0 & 72.4 \\
  & R-Zero             & 84.2 & 80.6 & 97.5 & 53.5 & 52.9   & 79.6   & 67.2 & 59.0 & 71.8 \\  
  & DAPO               & 74.0 & 65.4 & 97.5 & 48.8 & 45.8 & 68.3 & 64.6 & 54.0 & 64.8 \\
  & RLVE               & 81.9 & 75.0 & 95.0 & 51.5 & 49.6 & 75.8 & 67.7 & 57.3 & 69.2 \\
  & \textbf{\ours}     & 86.2 & \textbf{83.0} & \textbf{100.0} & \textbf{60.0} & \textbf{53.6} & \textbf{81.9} & \textbf{70.2} & \textbf{63.2} & \textbf{74.8} \\
\midrule
\multirow{5}{*}{\shortstack[l]{Nemotron\\Cascade-8B}}
  & Untrained          & 83.3 & 71.3 & 92.5 & 61.5 & 55.8 & 73.8 & 65.2 & \textbf{64.3} & 71.0 \\
  & R-Zero             & 83.3   & 71.5   & 97.5   & 55.8   & 50.1   & 73.3   & 58.1  & 62.1   & 69.0\\
  & DAPO               & 78.3& 63.5& \textbf{100.0} & 40.0& 40.6& 65.8& 61.6 & 60.6& 63.8\\
  & RLVE               & 79.8 & 69.4 & \textbf{100.0} & 55.4 & 54.3 & 74.2 & \textbf{66.2} & 56.6 & 69.5 \\
  & \textbf{\ours}     & \textbf{84.8} & \textbf{72.3} & \textbf{100.0} & \textbf{64.2} & \textbf{57.2} & \textbf{78.8} & 65.7 & 62.8 & \textbf{73.2} \\
\bottomrule
\end{tabular}%
}
\end{table}

Table~\ref{tab:main} presents the main results.
The rightmost column reports the mean pass@1 across the benchmarks in each row. Three patterns stand out.  First, \ours{} is the only method that improves all three model families.  It raises the average from $49.2$ to $53.1$ on Qwen3-4B-Instruct, from $72.4$ to $74.8$ on the already-strong Qwen3-4B-Thinking checkpoint, and from $71.0$ to $73.2$ on Nemotron-Cascade-8B, a thinking model from a different family.  These correspond to relative gains of $7.9\%$, $3.3\%$, and $3.1\%$, respectively. Second, the strong-model regime highlights the weakness of fixed curricula. On Qwen3-4B-Thinking, DAPO and RLVE reduce the average to $64.8$ and $69.2$, while \ours{} still improves it.  This suggests that, once a static verified distribution becomes saturated or misaligned, stable reward alone is not enough; the training environment must also evolve with the solver. Third, the gains are not confined to the exact environments used for training. \ours{} trains only on self-generated executable environments from ten seeds, without using problem-answer data from the evaluation benchmarks.  Nevertheless, the resulting policy improves on external benchmarks with different formats, including GPQA Diamond ($65.2\!\to\!70.2$) and LiveCodeBench ($59.0\!\to\!63.2$) on Qwen3-4B-Thinking. This suggests that the evolving environment pool induces transferable reasoning behavior rather than only memorizing synthetic environment templates.

\subsection{Training dynamics: lower training score can mean a healthier frontier}
\label{sec:exp:dynamics}
\begin{figure}[t]
\centering
\includegraphics[width=\textwidth]{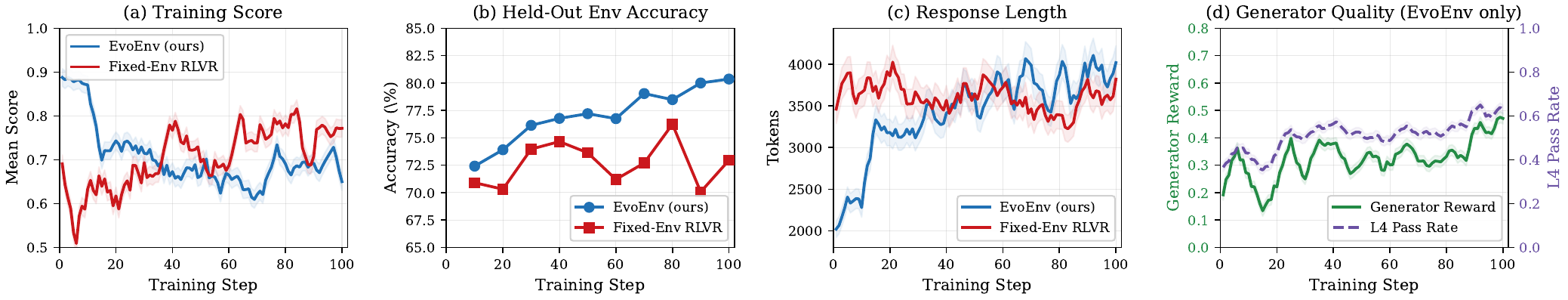}
\vspace{-10pt}
\caption{Training dynamics of \ours{} vs. fixed-environment RLVR on Qwen3-4B-Instruct-2507.  In \ours{}, the mean solver training score decreases from $0.88$ to $0.61$ as the generator synthesizes harder environments, while held-out accuracy on 50 unseen RLVE environments rises from $72.4\%$ to $80.4\%$~\cite{rlve}.  The fixed baseline stagnates near $72\%$.}
\label{fig:training_dynamics}
\vspace{-10pt}
\end{figure}

Figure~\ref{fig:training_dynamics} contrasts \ours{} with a fixed-environment RLVR baseline over 100 on-policy steps on Qwen3-4B-Instruct-2507.
The striking pattern here is the inverse relationship between training score and held-out accuracy.  In \ours{}, the mean training score decreases because the generator keeps raising the difficulty of the solver's environment pool.  Rather than indicating degradation, the declining score is the mechanism that preserves frontier signal: the solver is repeatedly placed in regimes where it can sometimes succeed and sometimes fail.  The fixed baseline maintains a roughly constant training score on a saturated pool and correspondingly shows little held-out improvement. The crucial effect is that environment synthesis keeps reward informative.

\subsection{Data-level audit: executable, expanding, and interpretable}
\label{sec:exp:ablation_data}

\begin{figure}[t]
\centering
\includegraphics[width=\linewidth]{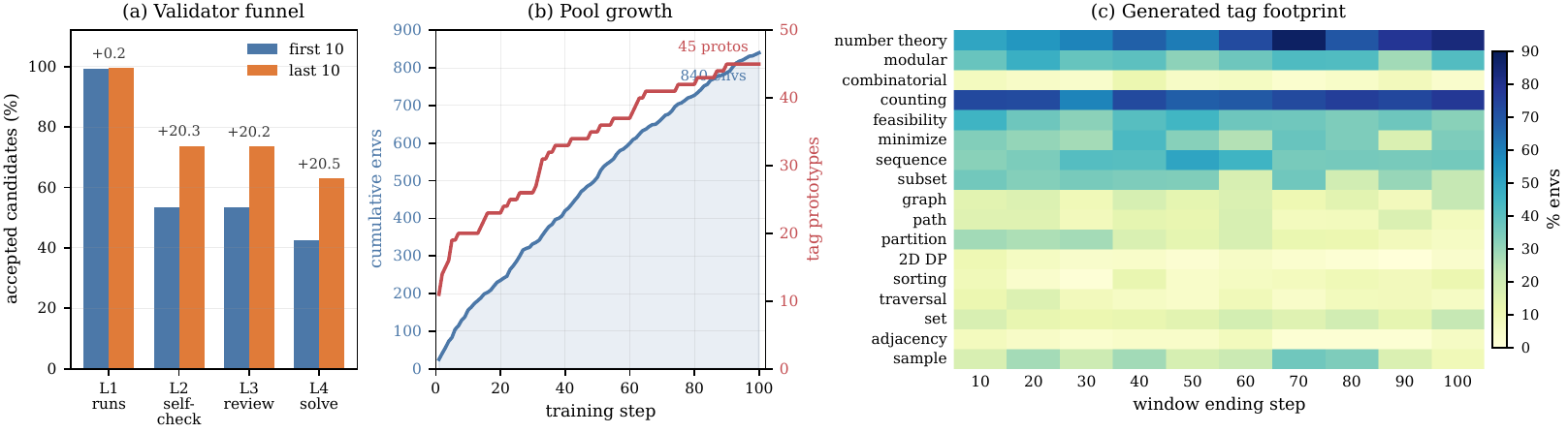}
\vspace{-10pt}
\caption{Data-level audit for the 100-step Qwen3-4B-Instruct run.  (a) The generator's accepted candidates improve through the validation funnel: L2/L3 self-check and review pass rates rise by about 20 percentage points, and L4 validation rises from $42.4\%$ to $63.0\%$ between the first and last ten steps.  (b) The accepted pool keeps growing, reaching 840 generated environments and 45 tag prototypes by step 100.  (c) Generated-only semantic tag frequencies across 10-step windows; the pool contains various types including counting, number-theoretic, modular, and sequence/subset-style environments rather than a single copied seed template.}
\label{fig:ablation_data}
\end{figure}

Figure~\ref{fig:ablation_data} gives three direct checks.  First, the generator becomes a better environment author under the actual admission pipeline: L2 and L3 pass rates increase from $53.4\%$ to about $73.6\%$, L4 validation increases from $42.4\%$ to $63.0\%$, and the generator reward rises from $0.292$ to $0.452$ between the first and last ten steps. 

Second, the accepted data stream does not collapse after the initial seed expansion.  Starting from only ten seed environments, \ours{} accumulates 840 accepted environments within 100 steps, while the tag-prototype count reaches 45.  Here a \emph{tag prototype} is registered whenever a new environment's binary tag vector has Jaccard similarity below $\tau{=}0.5$ to every existing prototype, so the count lower-bounds the number of structurally distinct environment families.  Reaching 45 prototypes from 10 seeds demonstrates that \ours{} generates environment types far beyond the initial seed distribution rather than recycling a narrow set of templates.

Third, the generated-only tag heatmap shows why these environments are useful for RLVR training.  The dominant mass is in countable, automatically checkable tasks---number theory appears in $68.2\%$ of generated environments, and modular or sequence structure in about $38\%$ each---but the pool also includes feasibility, minimization, subset, graph/path, traversal, set, and sampling mechanisms.  This gives the solver many executable reasoning problems with shared verifiability but varied surface forms, which is the data-side explanation for why \ours{} can keep supplying non-saturated reward.

\subsection{Reward component ablation: difficulty and diversity are jointly necessary}
\label{sec:exp:ablation_reward}
We isolate the contribution of the two generator-reward components: quality reward (validation and difficulty shaping) and diversity reward (novelty-gated exploration bonus).  Each ablation removes one component while keeping training steps, seeds, and evaluation protocol identical to the full \ours{} run on Qwen3-4B-Thinking-2507.
\begin{table}[t]
\centering
\caption{Reward component ablation on Qwen3-4B-Thinking-2507.  ``w/o Quality'' removes validation and difficulty shaping; ``w/o Diversity'' removes the novelty-gated exploration bonus.  Avg $\Delta$ is the mean improvement over the untrained baseline across the seven reported benchmarks.}
\label{tab:abl:reward}
\setlength{\tabcolsep}{4pt}
\small
\resizebox{\textwidth}{!}{%
\begin{tabular}{@{}l ccccccc c@{}}
\toprule
\textbf{Method} & \textbf{AIME'24} & \textbf{AIME'25} & \textbf{HMMT} & \textbf{B-AIME} & \textbf{Brumo} & \textbf{GPQA} & \textbf{LCB} & \textbf{Avg $\Delta$} \\
\midrule
Untrained           & \textbf{86.7} & 78.8 & 56.9 & 53.1 & 81.9 & 65.2 & 59.0 & --- \\
\textbf{\ours{} (full)} & 86.2 & \textbf{83.0} & \textbf{60.0} & \textbf{53.6} & 81.9 & \textbf{70.2} & \textbf{63.2} & \textbf{+2.4} \\
\quad w/o Quality   & 85.2 & 80.8 & 55.2 & 52.4 & \textbf{82.7} & 67.2 & 61.5 & +0.5 \\
\quad w/o Diversity & 85.2 & 81.5 & 57.3 & 53.4 & 81.2 & 66.7 & 60.4 & +0.6 \\
\bottomrule
\end{tabular}%
}
\end{table}
Both components are important.  Removing quality reward reduces the average gain from $+2.4$ to $+0.5$, with the largest drops on HMMT, GPQA, and AIME 2025.  Removing diversity reward reduces the gain to $+0.6$ and particularly hurts GPQA and LiveCodeBench, suggesting that template breadth is a driver of out-of-distribution transfer rather than merely a cosmetic regularizer.  The interaction is intuitive: difficulty without diversity yields narrow competence, while diversity without difficulty yields many environments that fail to supply gradient.

\section{Conclusion}
\label{sec:conclusion}

\ours{} suggests a path toward models that build part of their own training environment while retaining stable reward.  The key is not to trust the model's answers, but to ask it for executable structure that can be validated, frozen, sampled, and reused.  In zero-data reasoning RL, this turns self-improvement from a problem-generation loop into an environment-construction loop.  Compile-once, solve-many is the mechanism; the broader claim is that stable self-improvement may depend on models learning to construct the worlds that train them.

\newpage

\bibliographystyle{plainnat}
\bibliography{refs}

\newpage
\appendix
\section{Detailed positioning against nearby self-improvement methods}
\label{app:positioning}

Table~\ref{tab:positioning} expands the sketch in Section~\ref{sec:related} into a family-by-family comparison.  Two axes distinguish \ours{} from each adjacent family: the \emph{unit} that self-improvement synthesizes (one problem, one verifier, one trajectory, or a reusable environment), and the \emph{stability} of the reward source against further updates of the same policy.  A reward source that is policy-coupled, learned, or per-instance can still drive useful curricula, but it does not by itself give a durable, sample-many object that can be calibrated and reused.

\begin{table}[h]
\centering
\caption{Detailed positioning of \ours{} against nearby self-improvement methods.  ``Unit of synthesis'' is what the method produces and reuses across updates; ``reward stability'' summarizes whether the accepted reward depends on the current policy's beliefs; ``representative methods'' are illustrative rather than exhaustive.}
\label{tab:positioning}
\setlength{\tabcolsep}{3pt}
\footnotesize
\begin{tabular}{@{}p{0.13\linewidth}p{0.14\linewidth}p{0.14\linewidth}p{0.14\linewidth}p{0.18\linewidth}p{0.18\linewidth}@{}}
\toprule
\textbf{Family} & \textbf{Unit of synthesis} & \textbf{Reward / oracle source} & \textbf{Reward stability} & \textbf{Representative methods} & \textbf{Contrast with \ours{}} \\
\midrule
Self-questioning / self-instruction & One problem, question, or instruction & Majority vote, self-reward, or learned judge & Policy-coupled & R-Zero, SeRL, SQLM, TTRL, EMPO, Intuitor, OpenSIR~\citep{rzero,fang2025serl,chen2025selfquestioning,ttrl,empo,zhao2025intuitor,kwan2025opensir} & Adaptive curricula, but correctness moves with the learner. \\
Dual-play / co-evolution & Teacher, proposer, solver, or judge roles & Generated answers, judge feedback, or co-training signal & Mostly policy-coupled & PasoDoble, Socratic-Zero, MAE, EVOL-RL~\citep{zhang2025pasodoble,wang2025socraticzero,chen2025multiagentevolve,zhou2025evolrl} & Strong curriculum pressure; \ours{} externalizes accepted reward into executable artifacts. \\
Per-instance executable self-play & Program/input/output triple or per-task verifier & Python execution for the generated task or instance & Frozen per instance, consumed after one rollout & AZR, SCA, SPC, Agent0, SWE-Playground, SSR~\citep{absolutezero,zhou2025sca,chen2025spc,xia2025agent0,zhu2025sweplayground,wei2025ssr} & Grounded but per-instance; \ours{} validates and reuses environment-level oracles. \\
Corpus-grounded self-play & Document-grounded task & Retrieved-text grounding & Stable per corpus & SPICE, Search Self-Play~\citep{liu2025spice,lu2025searchselfplay} & Complementary grounding for knowledge tasks; \ours{} uses code grounding for algorithmic tasks. \\
Game / fixed-rule self-play & Trajectory in fixed rules & Game outcome or fixed simulator & Stable; rules are given & SPIRAL, Self-RedTeam, LSP~\citep{spiral,liu2025selfredteam,kuba2025lsp} & Rules are given; \ours{} synthesizes the rule-equivalent artifact. \\
Hand-crafted environment RLVR & Environment & Human-written scorer & Frozen per environment & RLVE~\citep{rlve} & Same training unit, but manually authored rather than self-synthesized. \\
Tool / agent environment synthesis & Tool sandbox and scenarios & Rule-based trajectory validation & Frozen post-pipeline & EnvScaler, AutoWebWorld, InfiniteWeb, VeriEnv, AutoPlay, EmboMatrix~\citep{song2026envscaler,wu2026autowebworld,zhang2026infiniteweb,chae2026verienv,ramrakhya2025autoplay,lei2025embomatrix} & Pipeline-built agent sandboxes; \ours{} is on-policy reasoning-environment co-evolution. \\
World-model simulators & Learned simulator state transitions & LLM as transition / world model & Depends on a learned simulator & DreamGym, UI-Simulator, WebEvolver~\citep{chen2025dreamgym,wang2025uisimulator,fang2025webevolver} & Reward depends on a learned simulator; \ours{}'s reward depends on Python execution. \\
Open-ended evolutionary & Agent or environment co-population & Population-level fitness & Varies across the population & POET, Darwin G\"odel Machine~\citep{poet,zhang2025darwingodel} & Broader open-ended agenda; \ours{} is solver-calibrated, on-policy, and validated. \\
Foundational self-improvement (lineage) & Generated rationales, self-generated preferences, or filtered traces & Verifier, learned reward model, or self-preference & Mostly policy-coupled & STaR, ReST-EM, SPIN, Self-Rewarding LMs, V-STaR~\citep{zelikman2022star,singh2024restem,chen2024spin,yuan2024selfrewarding,hosseini2024vstar} & Bootstraps from labelled seeds; \ours{} extends this lineage to environment-level objects. \\
\midrule
\textbf{\ours{} (this work)} & \textbf{Reusable verifiable environment} & \textbf{Frozen Python execution} & \textbf{Frozen between pool updates} & --- & --- \\
\bottomrule
\end{tabular}
\end{table}

\section{Anatomy of a verifiable environment}
\label{app:env_anatomy}

\begin{figure}[h]
\centering
\begin{tikzpicture}[
  font=\small,
  box/.style={draw, rounded corners=2pt, minimum height=0.7cm, minimum width=1.6cm, align=center, font=\footnotesize},
  codebox/.style={draw, rounded corners=2pt, fill=blue!5, minimum height=0.7cm, align=center, font=\footnotesize},
  llmbox/.style={draw, rounded corners=2pt, fill=orange!8, minimum height=0.7cm, align=center, font=\footnotesize},
  arr/.style={->, >=stealth, thick},
]
\node[codebox, minimum width=1.8cm] (seed) at (0,0) {seed $\sigma$, diff $\delta$};
\node[codebox, minimum width=2.4cm] (gen) at (3.2,0) {\texttt{\_generate}\\[-1pt]{\tiny sample $z$, compute $a^\star$}};
\node[codebox, minimum width=2.2cm] (prompt) at (6.6,0) {\texttt{\_prompt}\\[-1pt]{\tiny render NL task}};
\node[llmbox, minimum width=2.0cm] (solver) at (9.8,0) {\textbf{Solver LLM}\\[-1pt]{\tiny reason in NL}};
\node[codebox, minimum width=2.2cm] (parse) at (9.8,-1.4) {\texttt{\_process}\\[-1pt]{\tiny extract answer}};
\node[codebox, minimum width=2.2cm] (score) at (6.6,-1.4) {\texttt{scorer}\\[-1pt]{\tiny compare to $a^\star$}};
\node[box, minimum width=1.6cm, fill=green!10] (reward) at (3.2,-1.4) {reward $r$};
\draw[arr] (seed) -- (gen);
\draw[arr] (gen) -- (prompt);
\draw[arr] (prompt) -- (solver);
\draw[arr] (solver) -- (parse);
\draw[arr] (parse) -- (score);
\draw[arr] (score) -- (reward);
\draw[arr, dashed, gray] (gen.south) -- ++(0,-0.5) -| (score.north) node[pos=0.25, above, font=\tiny, gray] {$a^\star$};
\node[font=\tiny, blue!60!black, anchor=south] at (3.2,0.55) {\textit{deterministic execution}};
\node[font=\tiny, orange!70!black, anchor=south] at (9.8,0.55) {\textit{natural-language reasoning}};
\end{tikzpicture}
\caption{Anatomy of a verifiable environment.  Reference answers flow through the code path from generation to scoring; the solver sees only the rendered natural-language task.}
\label{fig:env_anatomy}
\end{figure}
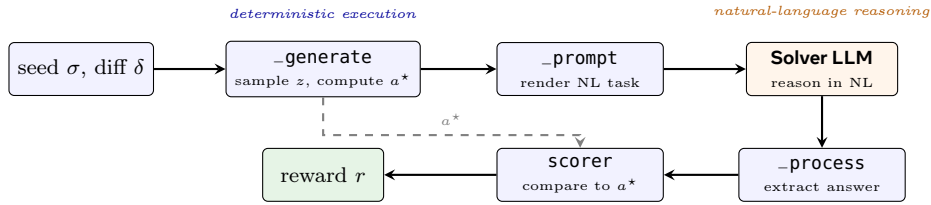

Figure~\ref{fig:env_anatomy} traces the data flow inside a single environment.  Reference answers are computed by deterministic Python execution and reach the scorer through the code path; the solver sees only the rendered natural-language prompt.  This is the picture that the validation pipeline (L1--L5 in Appendix~\ref{app:interface}) must protect: every safeguard ensures that one of these arrows continues to mean what it appears to mean.

\section{Additional interface and validator details}
\label{app:interface}
The main text describes each candidate as a subclass of \texttt{VerifiableEnvironment}.  Table~\ref{tab:validator_layers} summarizes the validation layers used before a candidate can enter the pool.  The layers are intentionally ordered from cheap syntactic checks to more expensive semantic and solver-facing tests, so early failures do not consume solver rollout budget.

\begin{table}[h]
\centering
\caption{Validator layers.}
\label{tab:validator_layers}
\setlength{\tabcolsep}{4pt}
\small
\begin{tabular}{@{}lp{0.73\linewidth}@{}}
\toprule
\textbf{Layer} & \textbf{Check} \\
\midrule
L1 & Parseable Python; expected class and methods exist. \\
L2 & Class instantiates; generation, prompt rendering, parsing, and scoring run on multiple seeds and difficulty values. \\
L3 & Determinism under identical seeds; prompt, state, and reference answer are stable. \\
L4 & Non-triviality across seeds and difficulty values; prompts and answers vary. \\
L5 & Scorer contract: reference answers score positively; perturbations, malformed answers, and type mismatches do not; parser does not leak hidden references. \\
\bottomrule
\end{tabular}
\end{table}

\section{Illustrative planted subset-sum environment}
\label{app:subset_example}
The following minimal feasibility-check example illustrates the environment interface.  The generator plants a solution and adds distractors.  The solver must find any valid subset, while the scorer checks the advertised constraint by verifying both the target sum and multiset multiplicities.

\begin{figure}[h]
\centering
\begin{minipage}{0.92\linewidth}
\footnotesize
\begin{verbatim}
from collections import Counter

class SubsetSumEnv(VerifiableEnvironment):
    prompt_template = ("Given the multiset {S}, find a nonempty "
                       "submultiset that sums exactly to {target}. "
                       "Output the elements.")

    def _generate(self, seed, difficulty):
        rng = Random(seed)
        n = 8 + difficulty * 4
        k = rng.randint(2, n // 2)
        solution = [rng.randint(1, 50) for _ in range(k)]
        self.target = sum(solution)
        self.S = solution + [rng.randint(1, 50) for _ in range(n - k)]
        rng.shuffle(self.S)
        self.ref_answer = sorted(solution)

    def _prompt_generate(self):
        return self.prompt_template.format(
            S=self.S, target=self.target)

    def _process(self, response):
        return sorted(parse_int_list(response))

    def scorer(self, parsed):
        if not parsed or sum(parsed) != self.target:
            return 0.0
        have, need = Counter(self.S), Counter(parsed)
        if any(need[x] > have[x] for x in need):
            return 0.0
        return 1.0
\end{verbatim}
\end{minipage}
\caption{A feasibility-check environment.  The stored reference answer is useful for validation, but the scorer accepts any subset satisfying the advertised constraint.}
\label{fig:subset_example}
\end{figure}

\section{Semantic-review audit against an external reviewer}
\label{app:semantic_audit}

The same-policy reviewer in Section~\ref{sec:method:validate} only gates pool admission; its verdict never enters the generator reward (Eq.~\eqref{eq:rgen}).  Even so, the filter would be cosmetic if it agreed with a stronger reviewer no better than chance, so we audit it offline.

\paragraph{Setup.}
We take $79$ environments that already cleared all five mechanical layers ($\ell(e)=5$) from earlier training runs.  Ground-truth labels come from a single review by GPT-5.4 ($35$ \texttt{has\_bugs}, $44$ \texttt{correct}); GPT-5.4 is not a perfect oracle but is a substantially more capable reasoner than our $4$B in-training policy.  The audited reviewer matches the deployed configuration: identical \texttt{Qwen3-4B Instruct-2507} weights, the same four-step structured prompt (data flow, instance trace, algorithm check, scorer check), $K_{\rm rev}=3$ samples at temperature $0.6$, aggregated by \emph{any-reject}.  

\paragraph{Agreement.}
Table~\ref{tab:semantic_audit} reports agreement under four aggregation rules.  The deployed any-reject configuration attains $\mathrm{F}_1=87.0\%$ ($\mathrm{P}=85.7\%$, $\mathrm{R}=88.2\%$, accuracy $88.6\%$), well above the always-\texttt{correct} baseline ($55.7\%$ accuracy, zero recall) and the always-\texttt{has\_bugs} baseline ($44\%$ precision). 

\begin{table}[h]
\centering
\caption{Agreement between the same-policy reviewer (Qwen3-4B-Instruct, $K_{\rm rev}=3$,
$T=0.6$) and GPT-5.4 on the $79$ environments with parseable verdicts ($35$ buggy,
$44$ correct under GPT-5.4).}
\label{tab:semantic_audit}
\setlength{\tabcolsep}{6pt}
\small
\begin{tabular}{@{}lccccc@{}}
\toprule
\textbf{Aggregation} & \textbf{Acc.} & \textbf{Prec.} & \textbf{Recall} & \textbf{F$_1$} & \textbf{(TP, FP, TN, FN)}\\
\midrule
$K=1$ (single review) & $65.8\%$ & $75.0\%$ & $34.3\%$ & $47.1\%$ & $(12,4,40,23)$\\
$K=3$, majority                    & $65.8\%$ & $78.6\%$ & $31.4\%$ & $44.9\%$ & $(11,3,41,24)$\\
$\mathbf{K=3,\ any\text{-}reject}$ & $\mathbf{88.6\%}$ & $\mathbf{85.7\%}$ & $\mathbf{88.2\%}$ & $\mathbf{87.0\%}$ & $\mathbf{(30,5,40,4)}$\\
$K=3$, all-reject                  & $63.3\%$ & $87.5\%$ & $20.0\%$ & $32.6\%$ & $(7,1,43,28)$\\
\bottomrule
\end{tabular}
\end{table}

\paragraph{Failure modes.}
A balanced manual case study matches the quantitative picture.  The reviewer reliably catches greedy-for-optimal substitutions, hardcoded or trivially-derivable reference answers, mathematical-property violations (e.g., a non-prime claimed prime), and most incorrect-recurrence bugs that surface on a small traced instance.  It is weakest on \emph{cross-method data flow} (a parameter written in \texttt{\_generate} but never surfaced by \texttt{\_prompt\_generate} is hard to spot when methods are reviewed in isolation), \emph{subtle data-generation pathologies} (e.g., randomly removing edges silently breaking connectivity assumptions), and \emph{hallucinated edge cases} on otherwise correct code.  These three modes account for most of the FN and FP in Table~\ref{tab:semantic_audit}.

\paragraph{Why this gate suffices.}
Residual disagreement does not directly contaminate the generator's gradient: the verdict enters only Eq.~\eqref{eq:accept}, not Eq.~\eqref{eq:rgen}.  A false-positive correct candidate still receives its mechanical $Q_{\rm val}+\gamma_t N_t$ reward and only loses its slot in the solver-training pool, while a false-negative buggy candidate is still filtered by the solver-relative pass-rate condition $0<\hat a_m<1$ before it can train the solver.  The reviewer therefore only needs to be biased correctly relative to the external reference, not to match it perfectly.

\section{Seed environments}
\label{app:seeds}
All runs start from the same ten initial seed environments.  The seeds are executable interface examples and initial pool items, not external problem--answer datasets: they define the file layout, parameter conventions, prompt-rendering conventions, and scorer conventions that the generator imitates, while leaving the generator to synthesize new environment code during training.  We deliberately picked the ten seeds to cover (i) different algorithmic categories, (ii) different oracle structures (exact match, element-wise match, partial-credit ratios, and feasibility), and (iii) different difficulty knobs (problem size, value range, structural density, recursion depth).  The exact ten-seed set is held fixed across all experiments. 

Table~\ref{tab:seed_detail} lists every seed individually.  Each row corresponds to one Python class implementing the \texttt{VerifiableEnvironment} interface from Appendix~\ref{app:env_anatomy}; the parameter knobs are the keys the generator fills in via \texttt{self.parameter} and the scorer column names the scorer family the seed exemplifies.  We expand on the four scorer families below the table.

\begin{table}[h]
\centering
\caption{The ten initial seed environments.  Each row gives the environment class
name, the algorithmic category it represents, the task asked of the solver, the
difficulty knobs surfaced to the generator (read from \texttt{self.parameter}),
and the family of scorer the seed exemplifies (defined below the table).}
\label{tab:seed_detail}
\setlength{\tabcolsep}{4pt}
\small
\resizebox{\textwidth}{!}{%
\begin{tabular}{@{}clllll@{}}
\toprule
\textbf{\#} & \textbf{Seed class} & \textbf{Category} & \textbf{Task} & \textbf{Difficulty knobs} & \textbf{Scorer family} \\
\midrule
1  & \texttt{Sorting}                     & Array / ordering             & Sort $N$ integers in ascending order.                                                              & $N$, value range                          & Element-wise \\
2  & \texttt{SlidingWindow}               & Array / deque                & Output the minimum of every contiguous size-$K$ window.                                            & $N$, $K$                                  & Element-wise \\
3  & \texttt{MonotonicStack}              & Array / stack                & For each $i$, count $j>i$ with $A[i]>\max(A[i{+}1..j])$, then sum.                                 & $N$                                       & Exact match \\
4  & \texttt{Knapsack}                    & 0/1 DP / optimization        & Pick items maximizing total value under a weight budget.                                           & $N$, $W_{\max}$, value range              & Partial-credit ratio \\
5  & \texttt{SubsetSum}                   & Combinatorial / feasibility  & Pick a subset of indices summing exactly to a planted target.                                      & $N$                                       & Feasibility \\
6  & \texttt{BoundedIntervalIntersection} & Combinatorial / counting     & Count non-empty interval subsets whose intersection has length $\ge K$.                            & $N$, $K$                                  & Partial-credit ratio \\
7  & \texttt{Bridge}                      & Graph / connectivity         & Find every bridge edge of an undirected graph.                                                     & $N$, component count, edge density        & Partial-credit ratio \\
8  & \texttt{EuclidGame}                  & Game theory / number theory  & Determine the optimal-play winner of the Euclid subtraction game on $(X,Y)$.                       & $\max(X,Y)$                               & Exact match (binary) \\
9  & \texttt{Fibonacci}                   & Linear recurrence            & Compute $A[N]\bmod m$ for $A[n]=P\cdot A[n{-}1]+Q\cdot A[n{-}2]$.                                  & $N$, modulus                              & Exact match \\
10 & \texttt{RecursiveFunction}           & Recursion / Ackermann-like   & Compute $f(M,N)$ under a three-case recurrence with self-nested calls.                              & $\max(M,N)$                               & Exact match \\
\bottomrule
\end{tabular}%
}
\end{table}

\paragraph{Why ten seeds suffice.} Despite their small number, the ten seeds expose the generator to every interface pattern used downstream.  They span seven algorithmic families, four scorer families, and structural primitives covering arrays, graphs, recurrences, optimal substructure (DP), and game-theoretic state.  Four seeds (\texttt{EuclidGame}, \texttt{Fibonacci}, \texttt{MonotonicStack}, \texttt{SlidingWindow}) are adapted from competitive-programming archives (Luogu OI); the remaining six are written from scratch to balance scorer-family and structural coverage.  Section~\ref{sec:exp:ablation_data} and Appendix~\ref{app:gallery} report on the trajectories the generator follows away from this starting set during training.

\section{Generated-environment data audit}
\label{app:gen_data}

\subsection{Qualitative examples within the 100-step run}
\label{app:gallery}
The examples below illustrate the same shift at the environment level.  We restrict this gallery to environments generated no later than step 100 so that it matches the ablation study in Section~\ref{sec:exp:ablation_data}.  Distances are Jaccard distances from the environment's tag set to the nearest one of the ten original seeds.

\begin{table}[h]
\centering
\caption{Representative generated environments from the 100-step run.  Prompts are shortened to their first clause for readability.}
\label{tab:qual_examples}
\setlength{\tabcolsep}{4pt}
\small
\resizebox{\textwidth}{!}{%
\begin{tabular}{@{}llrlp{0.46\linewidth}@{}}
\toprule
\textbf{Class} & \textbf{Step} & \textbf{$d_{\rm seed}$} & \textbf{\# tags} & \textbf{Prompt sketch} \\
\midrule
\texttt{PrimePartition} & 1 & 0.73 & 8 & Given a list of positive integers, count or decide prime-constrained partitions. \\
\texttt{PrimePairProductSum} & 10 & 0.71 & 6 & Given $N$ and a set of distinct prime numbers, optimize a product/sum expression. \\
\texttt{PrimeGCDChain} & 60 & 0.61 & 17 & Combine a sequence of values with GCD and prime-chain constraints. \\
\texttt{SubsetCycleGCD\_PrimeChain} & 69 & 0.65 & 16 & Fuse subset selection, cyclic structure, and GCD/prime constraints. \\
\texttt{CycleDistanceSum} & 83 & 0.75 & 9 & Reason over distances induced by a cycle-like combinatorial structure. \\
\bottomrule
\end{tabular}%
}
\end{table}

These examples are not meant to certify correctness by inspection; every accepted environment is still admitted through the validation and self-review pipeline described in Section~\ref{sec:method:validate}.  Their purpose is to make the distributional audit concrete: the generator moves quickly from the seed pool toward number-theoretic, modular, and sequence-structured problems that are solver-calibrated rather than trivially copied from RLVE.

\section{Limitations and Risks}
\label{sec:limitations}

\paragraph{Scope of the environment interface.}
This paper studies a controlled setting: zero-data RLVR for reasoning with deterministic Python environments.  The current interface is best matched to tasks where an executable sampler, oracle, renderer, and scorer can be written compactly and validated automatically.  This includes many algorithmic and feasibility-style tasks, but it does not establish the same claims for open-ended judgment, human preference modeling, long-horizon physical simulation, or interactive tool-use environments.  Extending verifiable environment synthesis to those settings would require different grounding mechanisms and stronger safety checks.

\paragraph{Executable-code safety.}
A method that asks a model to synthesize executable environments must treat generated code as untrusted.  Our implementation executes candidates in sandboxed subprocesses with restricted imports, resource limits, deterministic seeds, and wall-clock timeouts; candidates that violate the interface or execution constraints are rejected before pool admission.  These safeguards are part of the method rather than optional engineering details.  Any extension to richer tool-use, web, file-system, or embodied environments would require correspondingly stronger isolation and monitoring.

\paragraph{Broader impact.}
A positive implication of verifiable environment synthesis is that the training curriculum becomes more inspectable: accepted environments are concrete code artifacts rather than opaque pseudo-labels or hidden preference judgments.  This can reduce dependence on proprietary post-training datasets and make reward sources easier to audit.  At the same time, improving autonomous curriculum construction could amplify the capabilities of reasoning models in both beneficial and dual-use settings.  We therefore view transparent environment release, sandboxed execution, and careful auditing of generated artifacts as necessary safeguards for follow-up work.

\newpage
\section{Hyperparameters}
\label{app:hparams}

Table~\ref{tab:hparams_known} lists the headline hyperparameters reported in the
main text; Table~\ref{tab:hparams_full} reports the full configuration of the
deployed training run, grouped by component.

\begin{table}[h]
\centering
\caption{Training and evaluation hyperparameters reported in the main text.}
\label{tab:hparams_known}
\setlength{\tabcolsep}{5pt}
\small
\begin{tabular}{@{}ll@{}}
\toprule
\textbf{Hyperparameter} & \textbf{Value} \\
\midrule
Generator advantage scale $w_{\mathrm{gen}}$ & $0.3$ \\
Solver test budget for candidate calibration & $m=8$ \\
Semantic self-review samples & $K_{\rm rev}=3$ with any-reject aggregation \\
Prompt/code novelty mixture $\lambda$ & $0.5$ \\
Original-seed floor $\rho_{\min}$ & $0.2$ \\
Training steps & $100$ steps \\
Evaluation context & up to $128\mathrm{k}$ \\
Evaluation temperature / top-$p$ & $0.6$ / $0.95$ \\
\bottomrule
\end{tabular}
\end{table}
\begin{table}[h]
\centering
\caption{Full configuration of the deployed $4$B-Instruct training run.  Generator
group, solver group, novelty, pool, review, and sandbox settings are shared across
backbone variants; only the batch sizes and context lengths in the optimizer block
are rescaled when scaling up to $8$B.}
\label{tab:hparams_full}
\setlength{\tabcolsep}{5pt}
\tiny
\begin{tabular}{@{}llp{0.42\linewidth}@{}}
\toprule
\textbf{Block} & \textbf{Hyperparameter} & \textbf{Value} \\
\midrule
\multirow{4}{*}{Generator group}
 & Number of distinct generator prompts per step                & $16$ \\
 & GRPO group size $M$ (responses per generator prompt)         & $8$ \\
 & Generator sampling temperature                               & $1.0$ \\
 & Generator max response length                                & $8\mathrm{k}$ tokens (rollout side) \\
\midrule
\multirow{5}{*}{Solver group}
 & Solver batch size $B$ (\texttt{data.train\_batch\_size})     & $64$ \\
 & PPO mini-batch size                                          & $64$ \\
 & GRPO group size (\texttt{rollout.n}, responses per prompt)   & $8$ \\
 & Solver sampling temperature                                  & $1.0$ \\
 & Solver max prompt / response length                          & $8\mathrm{k}$ / $16\mathrm{k}$ tokens \\
\midrule
\multirow{6}{*}{Novelty schedule}
 & Novelty weight bounds $(\gamma_{\min},\gamma_{\max})$        & $(2.0,\,5.0)$ \\
 & Activation thresholds $(\tau_{\rm low},\tau_{\rm high})$     & $(0.45,\,0.65)$ \\
 & Pool-admission gate $\tau_{\rm gate}$                        & $0.80$ \\
 & EMA decay for $\bar s_t$                                     & $\alpha_{\rm ema}=0.6$ (init $0.5$) \\
 & Embedding model                                              & \texttt{all-MiniLM-L6-v2} (frozen) \\
 & Prompt/code mixture $\lambda$                                & $0.5$ \\
\midrule
\multirow{4}{*}{Calibration \& admission}
 & Calibration instances per L5 candidate $m$                   & up to $8$ (Eq.~\eqref{eq:empirical_accuracy}) \\
 & Solver responses per calibration instance                    & $1$ \\
\midrule
\multirow{4}{*}{Pool rotation}
 & Pool max epochs per environment                              & $5$ \\
 & Pool rotation cadence                                        & every $10$ training steps \\
 & Original-seed floor $\rho_{\min}$                            & $0.2$ \\
 & Initial pool                                                 & ten seeds (Appendix~\ref{app:seeds}) \\
\midrule
\multirow{4}{*}{Self-review}
 & Independent reviews $K_{\rm rev}$                            & $3$ \\
 & Aggregation rule                                             & any-reject \\
 & Review temperature                                           & $0.6$ \\
 & Review max response length                                   & $8\mathrm{k}$ tokens \\
\midrule
\multirow{8}{*}{RL objective \& optimizer}
 & Algorithm                                                    & GRPO  \\
 & PPO clip ratio (low / high)                                  & $0.2$ / $0.28$ \\
 & Dual-clip constant $c$                                       & $10.0$ \\
 & KL loss coefficient $\beta$                                  & $1\!\times\!10^{-3}$ \\
 & Entropy coefficient                                          & $0$ \\
 & Optimizer                                                    & AdamW \\
 & Learning rate / warmup / weight decay                        & $5\!\times\!10^{-7}$ / $5$ steps / $0.1$ \\
 & Gradient clip                                                & $1.0$ \\
\midrule
\multirow{4}{*}{Sandbox \& code}
 & Validation subprocess wall-clock timeout                     & $30\,\text{s}$ per layer \\
 & Allowed imports (enforced via generator prompt)              & \texttt{random}, \texttt{math}, \texttt{collections}, \texttt{itertools}, \texttt{heapq}, \texttt{bisect}, \texttt{functools}, \texttt{re}, \texttt{typing} \\
 & Disallowed imports                                           & any third-party (e.g.\ \texttt{numpy}, \texttt{networkx}, \texttt{scipy}) \\
 & Code extraction                                              & longest python fenced block \\
\bottomrule
\end{tabular}
\end{table}

\end{document}